# Arterial incident duration prediction using a bi-level framework of extreme gradient-tree boosting

Adriana-Simona Mihăiţă [13*], Zheyuan Liu [2], Chen Cai[1], Marian-Andrei Rizoiu[23]

[1] Advanced Data Analytics in Transport, DATA61|CSIRO, 2015, Sydney, Australia
[2] Australian National University, Canberra, 2601, ACT, Australia
[3] University of Technology in Sydney, NSW, Australia
*simona.mihaita@data61.csiro.au, adriana-simona.mihaita@uts.edu.au

**Abstract:** Predicting traffic incident duration is a major challenge for many traffic centres around the world. Most research studies focus on predicting the incident duration on motorways rather than arterial roads, due to a high network complexity and lack of data. In this paper we propose a bi-level framework for predicting the accident duration on arterial road networks in Sydney, based on operational requirements of incident clearance target which is less than 45 minutes. Using incident baseline information, we first deploy a classification method using various ensemble tree models in order to predict whether a new incident will be cleared in less than 45min or not. If the incident was classified as short-term, then various regression models are developed for predicting the actual incident duration in minutes by incorporating various traffic flow features. After outlier removal and intensive model hyper-parameter tuning through randomized search and cross-validation, we show that the extreme gradient boost approach outperformed all models, including the gradient-boosted decision-trees by almost 53%. Finally, we perform a feature importance evaluation for incident duration prediction and show that the best prediction results are obtained when leveraging the real-time traffic flow in vicinity road sections to the reported accident location.

## 1. Introduction
### 1. Context

Traffic congestion is a major concern for many cities around the world. Congestion arises due to various factors, including increased population, workforce concentration in central areas, or the lack of efficient public transport modes. Two forms of congestion are predominant: a) recurrent congestion during peak hours when traffic demand exceeds the road capacity, and b) non-recurrent congestion caused by unplanned events such as car accidents, breakdowns, weather, public manifestations etc. Previous studies have shown that almost 60% of traffic congestion is due to non-recurrent incidents with a stochastic behaviour in space and time [1]. In Australia, the number of road deaths per year was reduced by 70% since 1970s, however the annual economic cost of road crashes was estimated at $27 billion per annum in 2017 [2]. Traffic control centres have deployed various Traffic Incident Management Systems (TIMS) to handle the reported incidents, but the task is difficult due to various factors that can influence each incident, such as the number of vehicles involved in an accident, the time-of-day, the road characteristics, the traffic conditions etc. [3]. Accurately predicting the total duration shortly after an incident has just occurred could save not only operational costs but also lives. However, most prior work concentrates on predicting the duration of traffic incidents on freeways-motorways, a problem with well-defined traffic dynamics and often met with abundant data. In this study we employ a bi-level prediction framework using extreme boosted classification and regression models with the scope of predicting the incident duration on different arterial roads in the city of Sydney, Australia, with a focus on identifying as well the critical factors which impact the incident clearance times.

Initial methods used to predict the incident duration were mainly designed for inference and estimation, and they included: linear/non-parametric regression models [4], Bayesian classifiers [5], discrete choice models (DCM) [6], probabilistic distribution analyses [7], and the hazard-based duration models (HBDM) [8]. HBDMs are one of the most popular models due to their advantage of capturing the duration effects, accelerated failure times and easy integration with M5P trees models [9].

But more recently, new approaches in machine learning have emerged as a more apt way of predicting the incident duration due to their capacity to easily account for new data sources, as well as for removing the linearity assumptions between features and the predicted class [10]. Examples of such approaches are: artificial neural networks (ANNs) [11], genetic algorithms [12], support vector machines (SVM) [13], k-Nearest-Neighbours (kNN) [14] and decision-trees (DT)[15]. The recently proposed Gradient-Boosted-Decision-Trees (GBDT) have been shown to provide superior prediction performance compared to Random Forest, SVM and ANN [16]. However, it is known that GBDT can easily overfit when the prediction target has a long-tail distribution, as is the case of the traffic incident duration [16]. XGBoost [17] is another decision-tree enhancement method that has gained popularity recently in the machine learning community, due to its tree boosting capability, loss function regularization and adaptive learning rate. It was employed in several international competitions, winning 17 out of the 29 Kaggle competitions singled out on the 2015 Kaggle blog; it was also employed by every team in the top-10 in the 2015 KDDCup [18] for solving various problems such as store sales prediction, web text classification, hazard risk prediction, and product categorization. XGBoost's popularity is also due to its scalability (it can run on a single machine, as well as on distributed and parallelized clusters), its capacity to handle sparse data and the ability to handle instance weights in approximate tree learning (see recent paper published by [17] where authors proposed an end-to-end tree boosting system with cache-aware and sparsity learning features). While each of these methods have their own advantages and disadvantages, building a fast and reliable prediction framework which could be applied for real-time operations represents a true challenge.

## 2. Challenges and contribution

Most often, the accuracy of predicting the incident duration is not determined by the model which is being used, but rather by the learning methodology, the feature construction and the result interpretation. In this work, we address several open questions concerning the prediction of the traffic incident duration. First, the majority of prior work studied the prediction of incident duration on freeways or motorways, where the data accuracy is higher than on arterial roads; as of 2018 very few applied the prediction strategies on normal arterial roads due to high modelling complexity and location mis-matching; this is revealed by a recent state of art published in [19] which emphasises on the difficulty of solving this problem for arterial roads and the lack of studies in this field. Previous studies [20] also showed that the location of incidents decisively influences their clearance time, considering road characteristics and the difficulty for the rescue teams to clear the road. Incident prediction on freeways benefits further from the much faster association between the incident location and the freeway segment/traffic detector, and a simpler road structure with little ramifications. The first challenge that we try to solve is: can we accurately predict the incident duration on arterial/regular roads in a large city? Second, on many signalized road sections, consistent traffic information is hard to attain and integrate with a physical representation. For example, it is a challenge to map a reported incident location to a corresponding road section or a traffic detector if the reported location is either near a complicated intersection or far away from any road section with traffic information available. While it is known that traffic flow information can improve the accuracy of the prediction [12], there are not many studies which show how exactly to account for it, from what area, what road sections and from what timespan (before/after the incident was reported). Therefore, the second challenge that we are trying to solve is: how to construct a set of features which aggregate efficiently the traffic flow and the incident information, that are predictive of incident durations? More specifically, we want to detect what would be the most influential factors which affect the incident duration that traffic centres need to prioritise for a fast and efficient incident clearance.

In this paper we address both the aforementioned open questions. We deploy several machine learning models in a bi-level prediction framework combining classification and regression approaches; we use all current available and reported information with regards to the incident (location, severity, lanes affected, etc.), but also incorporate as well the traffic flow information collected from real loop detectors in the road network. The traffic flow information is also analysed through the use of 5 different feature sets which adds to the baseline incident information. To the best of our knowledge, this is the first research study employing a bi-level approach using extreme boost decision trees for incident duration prediction on arterial roads. Lastly, we analyse and rank the importance of each feature used in the prediction by calculating the Shapley values and show that the affected lanes, hour of the day when the incident happened, as well as the speed limit of the affected road section are the most important factors which influence the incident duration prediction. On the long-term, this work contributes to our ongoing objective to build a real-time platform for predicting traffic congestion in Sydney, and to analyse the incident impact during peak hours (see our previous works published in [21]-[22]). The paper is organized as follows: Section 2 presents the data sources available for this study, Section 3 showcases the methodology, Section 4 details the initial data profiling and feature correlation analysis, Section 5 presents the numerical results of the prediction framework, Section 6 details on the feature importance evaluation and Section 7 is reserved for conclusions and future perspectives.

## 2. Data sources
### 2.1 Baseline Feature construction

**Incident Data**: Our incident dataset covers one year (2017) of traffic incidents reported by the Traffic Management Centre (TMC) in Sydney. It contains 5,134 records of various planned and unplanned incidents that range from hazards, road closures to accidents and maintenance work. For this study, we focus on incidents labelled as "Accidents" since these induce the longest clearance time in the current subnetwork (according to TMC). There are 574 accident records in our dataset, with a mean duration of 44.59 minutes, and a maximum duration of 719 minutes. The arterial subnetwork covered by this dataset is represented in Fig 1 and expands over 3 main regional networks in Sydney: Ultimo, Rozelle and Ryde. We will refer to this area as the Victoria network in the rest of this paper. Fig 1 also shows the heat-map of the "accident duration" distribution in the Victoria network, which correlates with the hot-spots of the infrastructure where majority of long accidents tend to happen. This heat-map will be used to validate the findings on the most vulnerable road sections using the predictive model.

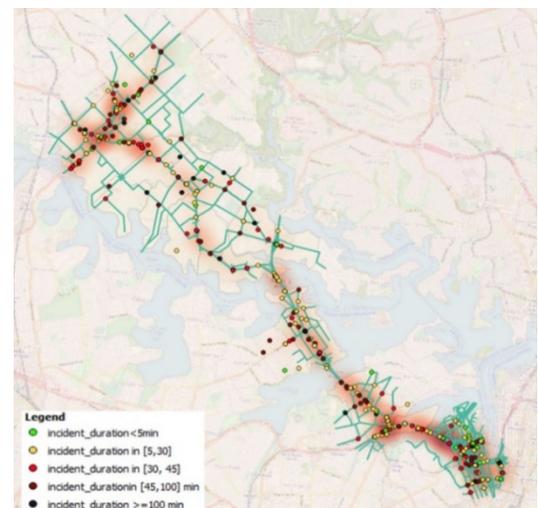

*Fig 1. Heat-map the "accident duration" distribution in the Victoria Rd subnetwork.*

Fig 2a) shows the distribution of the accident duration. The majority of accidents are cleared off in less than 30 min (291 out of 574), while the empirical complementary cumulative distribution function (ECCDF) in Fig 2b) reveals that the incident duration presents two different regimes given by different slopes to the right and left of a threshold T (identified to be revolving around 45minutes). We also observe that the incident duration is long-tail distributed, with the longest 10% of the incidents (57 out of 574) spanning between 100 and 719 minutes. This is likely to pose difficulties for prediction algorithms that assume linear dependencies. The original incident dataset holds comprehensive information for each incident such as location, description, number of lanes affected, direction, suburb etc. The list of the attributes that we employ for training our models is provided in Table 1, and it contains as well information about weather, events, area geometry and traffic flow data.

**Weather data** was extracted from the online database of the Australian Government Bureau of Meteorology and it accounts for the average temperature and rain for each day in 2017. We chose the observation station located in Northern Sydney (Observatory Hill) as the main source of weather data, as it is the closest station to the Victoria road network.

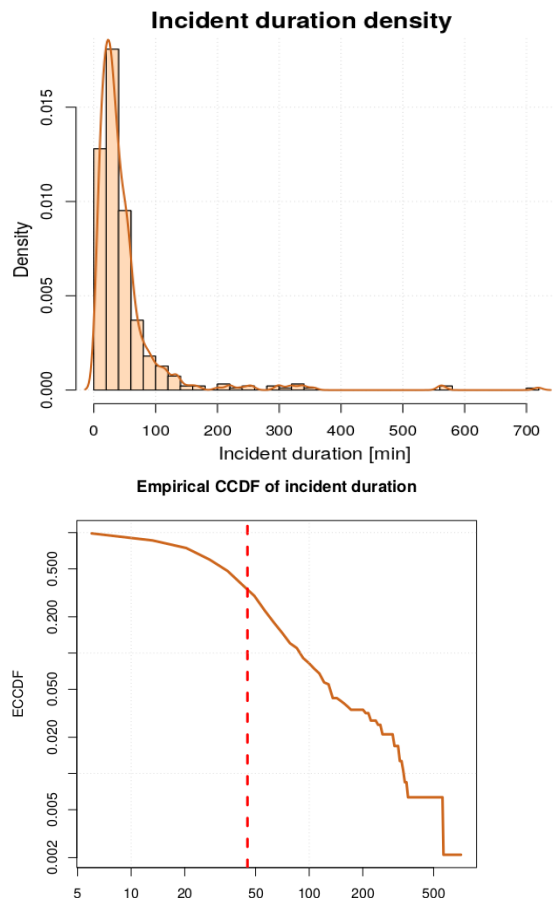

*Fig 2 a) Accident duration distribution and b) Empirical Cumulative Distribution Function of the accident duration.*

**Public Holiday features** were manually collected for 2017, which contains Boolean values for each day denoting whether the day is a public or regional holiday in New South Wales, Australia.

**Area geometry features** contain the sector ID as defined by TMC, the code of the official area where the accident occurred (as defined by the Bureau of Transport and Statistics), and supplementary information such as: section capacity, section speed limit, and number of lanes. These features are available for all road sections in the Victoria network, and they were extracted from the official traffic simulation model of the Victoria network, developed in Aimsun and previously used by the authors for incident impact analysis and traffic prediction [22].

### 2.2 Adding in traffic flow features

Traffic flow data is an important source of information as the traffic conditions on the affected road segment/s can contribute to the severity and impact of the reported incident. Traffic flow data is available for all the incidents in our dataset and it was extracted by first identifying all sections in the subnetwork which contain official counts from SCATS detectors. The real-time flow counts produced directly by the SCATS detectors are aggregated every 15-minutes. The data from each detector is summed according to their installation location to generate the flow data for the corresponding road section. Note that due to physical constraints, not all sections are equipped with SCATS detectors. There are 2,672 road sections in the model, 85 signalized intersections with the adaptive SCATS control system running, and a total of 4,256 SCATS detectors. Despite this large data traffic flow dataset, some incidents are reported in locations with no traffic flow information; hence the traffic flow feature associated to an incident without any close-by loop detectors has been replaced by N/A.

For our prediction, we use 3 measures for traffic flows: a) the reported real-time flow from the 15-minutes time-interval when the incident has been reported (TRF), b) the traffic flow corresponding to one hour prior to the accident (TFH) and c) the 15-minute to one hour traffic flow ratio on each section computed as TFR=TRF/TFH, in order to reflect the traffic evolution along the affected road section in the last one hour prior to the incident (a TFR close to zero means high congestion as the real-time flow decreases considerably close to the accident start-time as compared to the flow 60 minutes earlier).

A number of feature data sets (FS) are generated in order to: 1) investigate the effectiveness of the traffic information for incident duration prediction and 2) to detect the most affected road sections in the network from the feature importance evaluation later described in Section 5. These features sets are listed as follows:

**Baseline Feature Set (BFS)**: uses all the feature information from Table 1 apart from the traffic flow features; this feature set is used for the incident duration classification, but also as a baseline for predicting the actual incident duration.

**Feature Set A (FSA)**: adds features to the BFS to include flow counts from: a) all the existing road sections in the Victoria network both in real-time (TRFi); b) one hour prior to the accident (TFHi); and c) from the calculated ratio (TFRi, $i \in \{1, N_s\}$,), where $N_s$ is the total number of road sections). This resulted in the addition of almost 700 extra features in the model training.

**Table 1 List of features used in the prediction framework.**

| Categories | Features/explanation | Value dataset |
|---|---|---|
| Accident | Location | {X,Y} in GDA Lambert coordinates |
| | Hour of day | {0,1,…23} |
| | Peak Hour | {1,0} |
| | Day of week | {1,..5} |
| | Weekend | {0,1} |
| | Month of the Year | {1,2,..12} |
| | Type | {Accident} |
| | Subtype | {Bus, car, bicycle, animals, etc.} |
| | Affected lanes | {Null, 1 lane, 2 lanes, 3 lanes, 4 lanes, All lanes, breakdown} |
| | Direction | {East, West, North, South, E-W, N-S, One Direction, Both Directions} |
| | Severity | {1,2,..10} |
| | Incident Source | {1,2,3} |
| | Unplanned | {0-planned,1-unplanned} |
| Weather | Average Temperature | ranging from {11.13 ˚C – 32.4 ˚C } |
| | Rainfall | ranging from {0 – 85mm} |
| Events | Public holidays | {0-no,1-yes} |
| Area geometry | Sector ID | As defined by TMC |
| | TZName | As defined by BTS (Bureau of Transport Statistics) |
| | Section ID | $\mathbb{R}_+$ |
| | Section Speed | $\mathbb{R}_+$ [Km/h] |
| | Section Lanes | {0,1,2,3,4,5,6} |
| | Section Capacity | {0, max 3100 vehicles/hour} |
| | Section class | As defined by TMC |
| | Street ID | As defined by TMC |
| | Intersection ID | As defined by TMC |
| | Distance from CBD | $\mathbb{R}_+$ [Km] |
| Traffic data | TRF (Traffic flow in Real-Time) total number of vehicles detected by loop counts in the 15 min time interval when the accident happened | $\mathbb{R}_+$ |
| | TFH (Traffic flow historic): one hour previous to the accident. | $\mathbb{R}_+$ |
| | Traffic Flow Ratio TFR = TFR/TFH | Between 0 and 1 |

**Feature Set B (FSB)**: adds to the BFS the traffic flow extracted from only the top 5 closest road sections to the incident location (TRFi, TFHi, TFRi, where $i \in \{1,..5\}$). For incidents which are located in the vicinity of any official loop detectors/sections, we can obtain the corresponding traffic flow at which the incident occurred by identifying the nearest SCATS controlled road sections in the Euclidean distance. Note that if the incident has a geographic coordinate which is not directly adjacent to a SCATS-enabled section, the radius of the search area is expanded to encompass the 5 closest road sections with SCATS detectors.

**Feature Set C (FSC)**: adds to the features in BFS the aggregated traffic flow from the top 5 closest road sections ($TRF_{top\_5} = \sum_{i=1}^{5} TRF_i, TFH_{top\_5} = \sum_{i=1}^{5} TFH_i, TFR_{top\_5} = \sum_{i=1}^{5} TRF_i$). This feature set is a summary version of FSB intended to decrease the computational time and reduce the total number of features used to train the models.

**Feature Set D (FSD)**: adds to the BFS the traffic flow extracted from all the sections in the vicinity of the reported location of the incidents ($TRF_{dv} = \sum_{i=1}^{N_r} TRF_i$, $TFH_{dv} = \sum_{i=1}^{N_r} TFH_i$, $TFR_{dv} = \sum_{i=1}^{N_r} TRF_i$, where $N_r$ is the total number of road sections in the selected area and $dv$ is the distance from the location of the incident to the closest road sections). The main difference between the feature set D and C occurs in the description segments in dense areas: while FSC will only account for the aggregated traffic on closest 5 segments, here we account for all segments inside a vicinity area with a $dv$ radius, regardless of their number. We aggregate all traffic flow found in the vicinity area due to a high variability of section results for each reported location; for example, from one incident to another the total number of traffic features extracted through this method can vary a lot, depending mainly on the location of the accident. For a better evaluation of this feature set we also construct a sensitivity analysis and evaluate the performance of the prediction for $dv$ ranging in the following range: {100m, 200m, 300m, 500m, 600m}. While $dv$ could be further extended, from our previous analysis and data profiling we observed that the influence of traffic accidents along the selected corridor can have a major impact on the surrounding traffic up to 600m.

## 3. Methodology

Clearing any reported accidents in a short time represents a high priority for the traffic management centres around the world. In New South Wales, Australia, the target clearance time for traffic incidents is set to 45 minutes. Therefore, in the rest of this paper we will refer to any incidents cleared in less than 45min as "short-term incidents", and to those taking longer as "long-term incident".

The methodology we propose for modelling the duration of these incidents is based on a bi-level prediction framework combining a classification and regression modelling, which is represented in Fig 3. This approach has been constructed by considering the real-time operational goals of TMC, and with the objective of providing fast prediction into the lifecycle of incident clearance. Based on the initial traffic incident information, the first step is the deployment of a fast classification method which would only predict whether the accident will be cleared off in 45 minutes or not (Step 1). If the incident has been classified as "short" then an intelligent regression method will be called in order to estimate exactly the duration of the incident in minutes (Step 2). This step requires not only the baseline incident information, but also the information about the latest traffic flow in the vicinity area to the accident location (as previously described in Section 2) in order to improve the prediction accuracy. In the case where the classification predicted that the accident will take longer than 45min to be cleared off, this raises a higher challenge and more extensive features might be needed to improve the prediction accuracy (Step 3), such as: a) longer-term traffic flow information (both real and historical patterns), b) features extracted from the incident description of operators (through natural language processing methods for example), and possibly a c) a graph congestion propagation modelling as well.

In this paper we only present the construction of the bi-level modelling approach for predicting the short term-incident duration (Step 1 and 2) and leave the long-term incident duration problem (Step 3) for a future extension of our work. Various models have been applied for both the classification and the regression problem, with a special emphasis on the extreme boosted tree modelling which are detailed in the following sections.

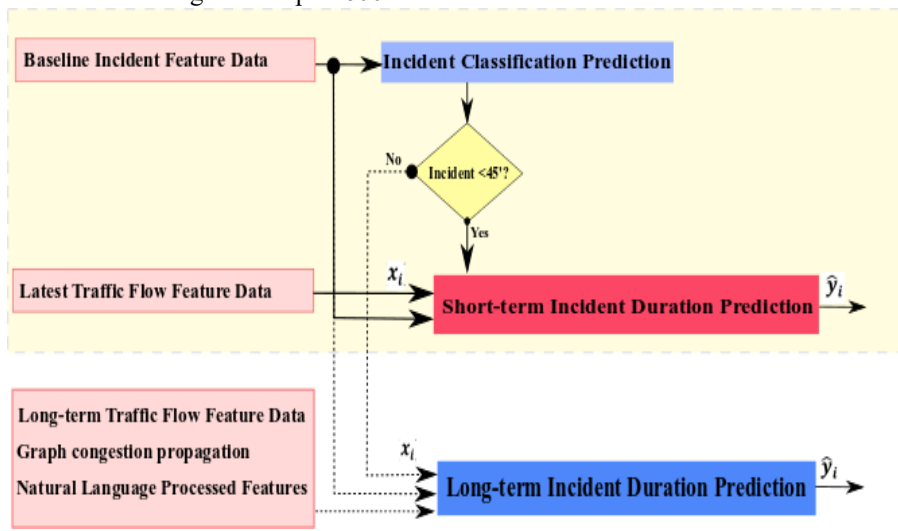

*Fig 3 Bi-level incident prediction framework.*

### 3.1 Incident duration classification

In the first step of the bi-level prediction framework, we aim to predict whether the accident duration will be lower or greater than 45min, by using a vector of descriptive features constructed from latest reported incident information. We denote as $X = [x_{i,j}]_{i=1,..N_i}^{j=1,..N_f}$ the matrix of model features, where $N_i$ is the total number of incidents used for training the models, and $N_f$ is the total number of features to be considered. We also denote the incident duration vector as $Y = [y_i]_{i=1,..N_i}$, where $y_i$ is the duration (in minutes) of an incident occurring at a specific time. The classification problem is to predict $Y$ from $X$, more specifically predict if $\hat{y}_i$ takes one of the following values:

$$\hat{y}_i = \begin{cases} 1, & if \ y_i \leq \bar{Y} \\ 0, & if \ y_i > \bar{Y} \end{cases} \quad (1)$$

where $\bar{Y}$ represents the 45min incident clearance time.

To summarize, given an incident record with a feature vector $x_i$ - as the $i^{th}$ row in the feature matrix $X$, the objective of the classification is to predict whether its duration $y_i$ is higher or lower than the expected clearance time $\bar{Y}$. For solving this classification problem, we have implemented and hyper-tuned various models which have been shown to provide good accuracy for similar topics, such as k-nearest neighbours (kNN)[14] which doesn't require specific assumptions about the data distribution or characteristics of the variables to learn, logistic regression (LR) [13] which focuses on the conditional probability distribution of the predicted variable given its set of features, random forests (RF) [23] which construct a multitude of decision trees during the training process where the leaves represent the final predicted class, gradient-boosted decision trees (GBDT) [16] and extreme-boosted decision trees (XGBoost) [17]. The latter two are powerful enhancements of decision trees which can be used for both classification and regression problems and are further detailed in the next section).

**Performance metrics**: to quantify both the prediction performance and the confidence level in the classification prediction, we perform a five-fold cross-validation (5CV). The dataset is randomly divided into five subsets (or folds), each containing the same proportion of the positive and the negative class (e.g. stratified folds). Iteratively, each fold serves as a test set, while the remaining four folds are used as training set. The model parameters are fit on the training set, and the predictions of the incident durations are obtained for the test set. We evaluate and present the mean prediction performances over the five folds using four widely used measures: Accuracy (A), Precision (P), Recall (R) and F1 which are calculated based on the True Positive (TP), True Negative (TN), false negative (FN) and false positive (FP) results as follows:

$$A = \frac{TP+TN}{TP+TN+FP+FN}, P = \frac{TP}{TP+FP},$$
$$R = \frac{TP}{TP+FN}, F_1 = \frac{2PR}{P+R} \quad (2)$$

Accuracy is the ratio of correct predictions over all predictions, but it is known to be very sensitive to class imbalance. Precision measures how many of the predictions made by the learner are correct, while Recall measures how many of the correct (e.g. true) examples were correctly predicted by the learner. The F1 measure is the geometric mean of Precision and Recall. In other words, to maximize F1, the learner needs to have simultaneously a high Precision and a high Recall which is the best performance metric reflecting how well the classification model is performing. The performance of each model against the above metrics is further presented and discussed in Section 5.

### 3.2 Incident duration prediction using regression

In the second step of the bi-level prediction framework, we aim to predict the exact incident duration in minutes if at the previous step the incident has been classified as "short-term". For this, we implement various models of regression, in which the continuous target variable needs to be predicted using extended features. Two of the most widely used and performant regressors are the GBDT and the XGBoost, which are ensemble tree methods making predictions using a tree-like structure. In a decision tree, each node makes a decision: the leaves are the final decisions (i.e. the predictions), while the non-leaf nodes guide the decision process towards the leaves. The path from the root to a leaf can be interpreted as a decision path, a conjunction of the decisions at each node. Consequently, the tree can handle non-linear interactions between features and the response (class) variable. GBDT is an enhanced version of decision trees which sequentially processes a combination of trees from weighted training data with a slow learning rate (25). This feature allows GBDT to handle unbalanced incident duration data. However, they can easily over-fit and require tree reconstruction whenever new data becomes available.

XGBoost enhances GBDT by: a) introducing a regularization parameter in the learning objective function (to control over-fitting), b) using sparsity-aware algorithms for parallel tree learning and c) by having a better support for multicore processing which reduces computational time. We further define the XGBoost model as follows. Given a dataset $D = \{(x_{ij}, y_i)\}$, where $|D| = N_i$ and $\{x_{ij} \in \mathbb{R}^{N_f}, y_i \in \mathbb{R}\}$, the XGBoost model uses $K$ additive functions to predict the incident duration as:

$$\hat{y}_i = \emptyset(x_{ij}) = \sum_{k=1}^{K} f_k(x_i),$$
$$f_k \in \mathcal{F} \quad (3)$$

where $K$ is the number of generated trees, and $f_k$ are functions in the functional space $\mathcal{F}$ defined as:

$$f_k(x_i) = \omega_{q(x)}, \omega \in \mathbb{R}^T, q: \mathbb{R}^{N_f} \to \{1,2,..T\},$$

$T$ is the number of leaves in the tree; each $f_k$ corresponds to an independent tree structure $q$ and leaf weights $\omega_{q(x)}$. The objective function to be minimized is given by:

$$\mathcal{L}(\emptyset) = \sum_{k=1}^{K} l(y_i, \hat{y}_i) + \sum_k \Omega(f_k) \quad (4)$$

where $l(y_i, \hat{y}_i)$ is the MAPE function defined later in Eq. (6) and $\Omega$ is the regularization term which penalizes the complexity of the model, and is expressed as:

$$\Omega(f_k) = \lambda T + \frac{\gamma}{2} \|\omega\|^2 \quad (5)$$

This additional regularization parameter helps to smooth out the learned weights and to avoid overfitting; if is set to zero then the objective problem falls back into a GBDT problem. For this type of models is impossible to enumerate all the tree structures that could be built and therefore greedy algorithms are used to start from leaves and iteratively add branches to the tree. Other techniques that can be used to prevent overfitting for this model are the shrinkage and column subsampling [17].

**Performance metrics**: the metrics defined in Eq. (2) cannot be used to measure the performance of an incident duration prediction through regression, which is usually quantified by the deviation of the prediction from the true target variable values. Among the most popular deviation measures, in this work we employ: a) the Mean Square Error (MSE), b) the $R^2$ determination coefficient and c) the Mean Absolute Percentile Error (MAPE). MSE can be sensitive to outliers and a long-tailed target variable. Consequently, in this work we finally measure the model performance using MAPE and $R^2$ defined as:

$$MAPE = \frac{100\%}{n} \sum_{i=1}^{n} \left|\frac{y_i - \widehat{y_i}}{y_i}\right| \quad (6)$$

$$R^2 = 1 - \frac{\sum_{i=1}^{n}(y_i - \widehat{y_i})^2}{\sum_{i=1}^{n}(y_i - \overline{y})^2} \quad (7)$$

In the results part of the paper, we present the comparative results of all models' performance, under different training and validation settings, with various loss functions and hyper-parameter tuning, before converging towards the most adapted model for solving our problem.

### 3.3 Hyper-parameter tuning through randomized search

Most machine learning algorithms have a set of hyper-parameters – parameters related to the internal design of the algorithm that cannot be fit from the training data. Both GBDT and XGBoost feature dozens of hyper-parameters, out of which the most important ones are considered to be *max_depth, learning_rate, min_child_weight, gamma, subsample, colsample_bytree and scale_pos_weight* [24]. The model hyper-parameters are usually tuned through search and cross-validation. The most extensive search technique is grid-search, in which several equally spaced points are chosen in the most likely interval for each parameter, and for each, a model is fitted and tested through cross-validation. The grid-search parameter tuning is straightforward; however, grid-search scales badly as the number of hyper-parameters increases. In this work, we employ a Randomized-Search [25] which selects randomly a (small) number of hyper-parameter configurations to use through cross-validation. It has been shown that for a high enough number of random samples (e.g. 100-200) the random search produces results comparable with the grid-search, in only a fraction of the time required by the grid-search. In our own experiments, randomized search has outperformed another popular heuristic, i.e. Bayesian optimisation.

For both classification and regression, we tune the hyper-parameters on each training data set, at each learning fold using 500 random combinations, evaluated using a 5 cross-validation. For the 10 cross-validation setup we used, we train each learner for 10 times (number of learning folds) x 5 times (number of hyper-parameter tuning folds) x 500 (number of random hyper-parameter combinations) = 25,000 times. This translates in a total execution time around 10 minutes, on a computational machine with 24 cores.

### 3.4 Evaluating the importance of most influential features

Generally, the influence of each feature data set can be very different on the prediction results and not all features are efficient for improving the result accuracy. In order to evaluate the effectiveness of these features we perform an evaluation of the most influential features. In contrast to the general modelling approaches such as support vector machines, linear regression or the autoregressive integrating moving averages, the Gradient Boosted Decision Trees and the Extreme Boosted Decision Trees are capable of ranking the influence of each feature on the predicted variable through different evaluation measures. For example, Ma et all. [16] proposed an aggregated measure on the ensemble of additive trees which used the summation measure originally developed by Breiman [26] (this summation measure is calculated over the terminal nodes which takes into consideration the improvement in the form of squared error as a result of using a predictor as a splitting variable in a non-terminal node). But for large sets of features entailing longer and complicated decision tree construction, we would need a fast metric for feature evaluation, which would allow us to drop easily any unnecessary information from the dataset. Our initial Baseline Feature Set contains 26 features, but with available traffic flow information on each road section in the incident vicinity (see explanation of features sets B, C and D), the dimension of the feature set would vary for each incident location and could easily reach hundreds of new added features in order to capture the impact of surrounding traffic on the clearance time.

Therefore, for an efficient evaluation of the feature importance through the proposed models, we use the Shapley value which originates in the game theory originally developed by Lloyd Shapley in [27]. The setup represents a coalition *S* of *n* players (belonging to a set *N)* that cooperate to obtain an overall gain, called the worth of the coalition *v(S)*. Since some players may contribute more to the coalition than other, the main question to be answered is what final distribution of the general surplus should be assigned to each player? Or more specifically to our problem, how to determine which feature is more important to the prediction result than others?

Therefore, the Shapley value is a fair distribution of gain among the players, assuming they all cooperate. For our prediction *p*, the Shapley value for a specific feature *i* (out of the total $N_f$ features) can be expressed as:

$$\emptyset_i(p) = \sum_{S \subseteq N\{i\}} \frac{|S|!(|N|-|S|-1)!}{|N|!}(v(S \cup \{i\}) - v(S)) \quad (8)$$

where $v(S \cup \{i\}) - v(S)$ represents what would the prediction of the model be without a specific feature *i*. Practically the Shap value calculates the marginal contribution of a feature *i* to the entire feature set over the number of features excluding *i*. The results of the feature importance evaluation will be further discussed in the section of results.

### 4. Data profiling

The initial data investigation around the available features revealed that some incident records are reported with very low durations (such as 0 or 1 minute). In reality, rarely accidents have such low durations, therefore we label these records as erroneous (or outliers). Outliers add noise in the training of the model, which may reduce the overall model performance [28]. As a consequence, the outliers are usually detected and filtered from the train set before the learning process. In Section 4 we discuss how the model performance improves when outliers are incrementally detected and removed from the model training.

Another data profiling investigation was to understand the correlation between the features in the data sets in order to detect whether they influence each other or not. Fig 4 showcases the Pearson correlation plot for all features which reveal, for example, that features which describe the network geometry are highly correlated (section ID, section speed, number of lanes and section capacity); this is mostly natural as network geometry surrounding the incident location play a similar importance in how long it will take before the incident get cleared off. An interesting finding is that the severity of a reported accident seems to be strongly positively correlated with the number of affected lanes (0.95) and the direction of travel (+0.59). This is mostly tied to the congestion accumulation along urban areas when all lanes in all directions are blocked for example, which makes it very hard for rescue team/police/firemen to access the location. The last observation is mostly related to the correlation between the distance from CBD and the associated urban zone name where the accident took place; once again this is a physical geometrical correlation. The rest of features do not present strong correlation with each other nor with the predicted target variable which indicates that the problem we are trying to solve is difficult and the dataset might represent a limitation in achieving excellent performance metrics, especially in the regression study. In the following we present our results for the bi-level approach and discuss the outcomes and possible improvements.

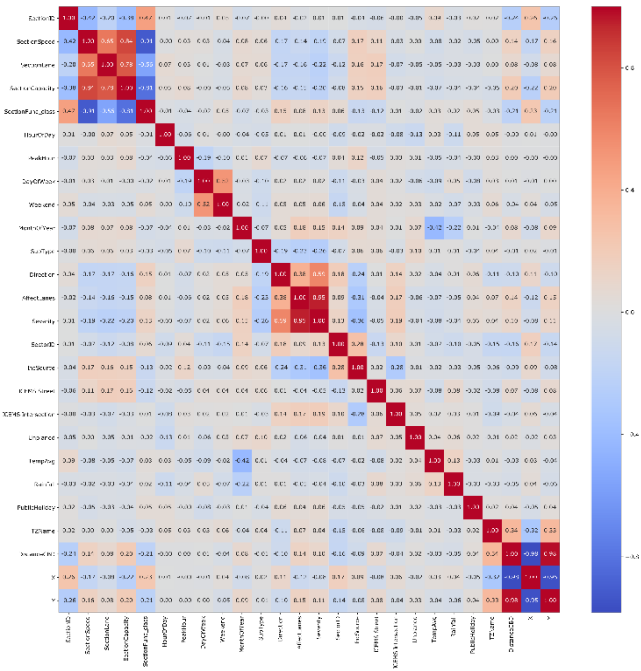

*Fig 4 Feature correlation plot.*

## 5. Numerical Results
### 4.1 Incident Classification results using BFS

The first step in the proposed bi-level prediction approach is to use only the existing accident information from the baseline feature set and determine whether a new accident will last less than 45 minutes or not. As previously mentioned in sub-section 3.3, for solving the classification problem we tune the model hyper-parameters using a 5 cross-validation and randomized-search using F1 (Eq. (2)) as the loss function.

For each fold, we compute the mean and standard deviation for Accuracy, F1, Precision and Recall.

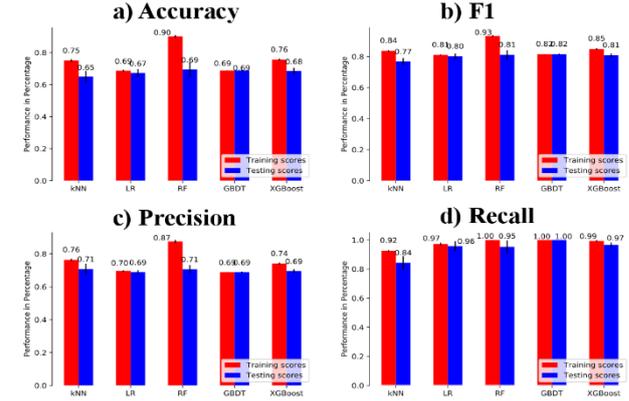

*Fig 5 Performance comparison of different classification algorithms on the Baseline Feature Set: a) Accuracy b) F1 c) Precision d) Recall.*

Fig 5 shows the performance of each of the five methods introduced in Section 3. After hyper-parameter tuning, all methods yield a testing accuracy above 65%; kNN ranks lowest at 65%, and RF and GBDT sharing the highest score of 69% (Fig 5a)). The training accuracy of RF is 21% higher than its testing score, indicating a sign of overfitting; in contrast, the remaining four ML models do not share such a characteristic. The F1 comparison between the five models reveals similar properties as the accuracy comparison (Fig 5b)). The highest F1 score in the testing (82%) is achieved by GBDT. Aside from kNN, all other four ML models have a testing Recall score above 95% (Fig 5d)), indicating that they are highly capable of identifying incidents that are shorter than 45 minutes. The Precision also shows that the tree-based methods – RF, GBDT and XGBoost – achieve the highest performance across all methods (Fig 5c)). Even though their testing performance are similar, RF appears to over-fit the training data more than the other two models.

The high accuracy results obtained in the classification part provides confidence that applying decision trees with extreme feature boost enhancements outperforms the majority of other methods and provides a good classification for majority of newly reported accidents.

### 4.2 Incident duration prediction using regression

The second part of the bi-level prediction problem focuses on predicting the actual incident duration in time and is more challenging as it requires additional feature information data especially on the traffic condition at the moment of the incident. Similarly to the classification, we tune the hyper-parameters using randomized-search; for each fold we compute MAPE and $R^2$, and we report their mean and standard deviation. We first use the most powerful models performing regressors GBDT and XGboost which we initially trained using the baseline feature data set. We have also evaluated both model's performance with and without the identified outliers and conducted various tests of the model performance when removing: a) zero-duration incidents, b) 1-minute duration incidents, etc.

Our analysis identified there are 27 (out of 574) incidents with a duration lower than 5 minutes. Removing these outliers reduced the MAPE from 221.04 to 120.34 for GBDT (see *Fig* 6a)), and from 77.84 to 68.77 for XGBoost (see *Fig 6*b)). Given the long-tail distribution of incident duration, we also train the methods in log-space which is known to reduce the impact of outliers, and retransform the final predictions for validation. For GBDT this procedure reduces the error considerably (from 120.34 to 83.26 after outlier removal), however this has limited impact for the extreme boost model most likely due to its additional regularization terms. Given these initial findings, all the modelling results presented in the following have a learning in the original incident duration space and are trained without the previously mentioned outliers.

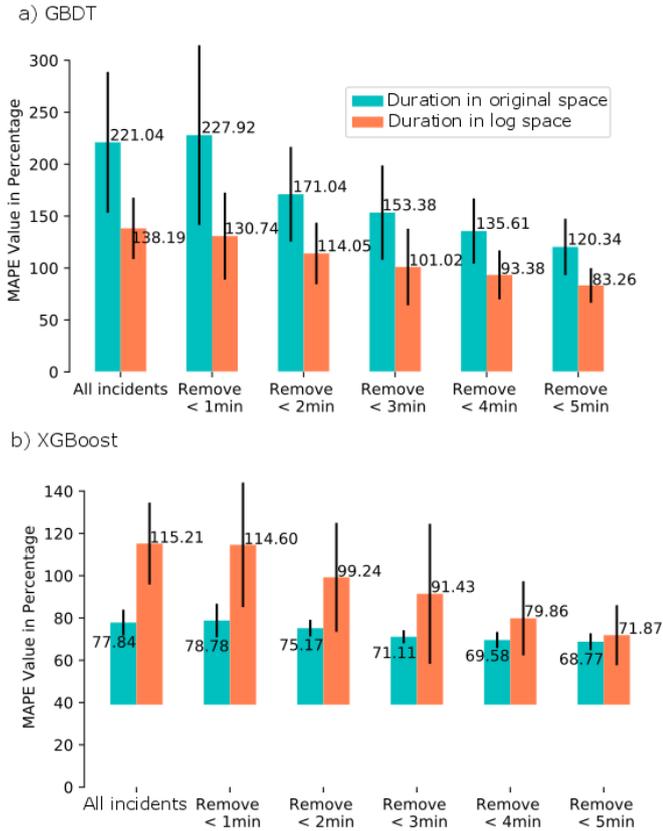

*Fig 6 MAPE comparison among datasets with incremental outliers removed.*

**Model comparison and fitting:** Going further, we evaluate the performance of four different regression methods which have shown good performance in our prior work: GBDT, XGBoost, RF and Linear Regression with Ridge Regularization (LR). Not only we evaluate their performance using two performance metrics, but we also train them with different loss functions such as MAPE, MAPE, MSE and $R^2$. *Fig 7*a) presents the performances of each regressor – measured by MAPE – and indicates that for all training loss functions, the XGBoost method considerably outperforms all other methods. The best XGBoost performance has a MAPE of 68.77, while RF – the second-best performer – achieves an almost double error (MAPE=117.2). We also do make the observation that our results are obtained on a complex urban network setting in urban areas, which renders the prediction exercise inherently more difficult than for motorways, where the road network structure is much simpler. Furthermore, our incident mapping is based on reported locations which are often far from official road sections, therefore any speed/lane association to real-life road structure becomes quite challenging. This represents a limitation which could be further enhanced by adding in a graph congestion propagation research study, a future extension of our work.

*Fig 7*b) reports the performances of the four methods using the $R^2$ coefficient as a goodness-of-fit measure. Similar conclusions can be drawn, with XGBoost outperforming all other methods in all training loss functions ($R^2$=0.78). The best result is obtained using MAPE as a training loss function. Consequently, we select XGBoost trained with MAPE as the best performing method, as we use it in for evaluating the incident duration prediction when traffic flow data is available.

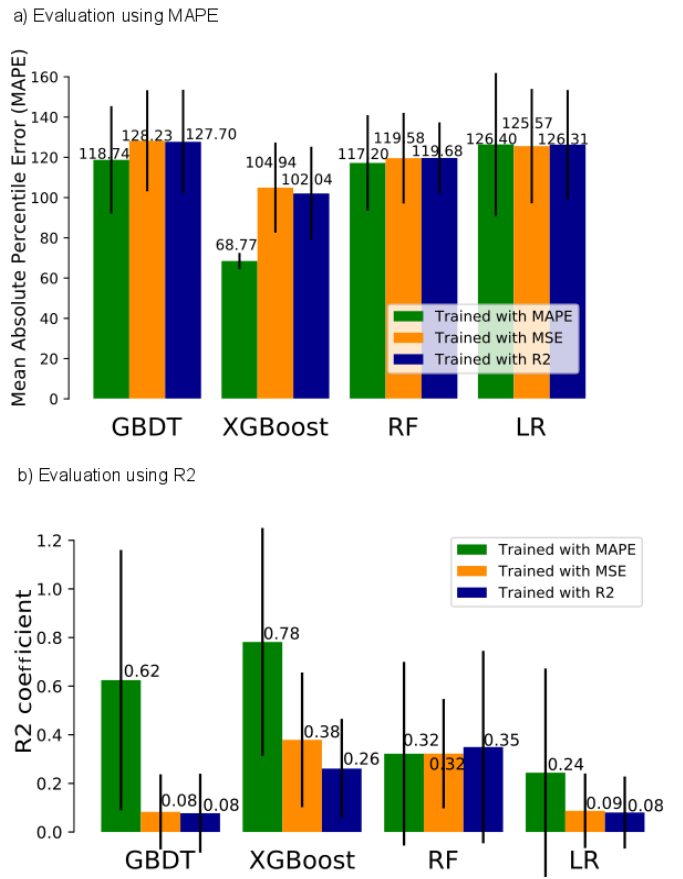

*Fig 7 The performances of four regressors are compared using two performances indicators: a) MAPE and b) R2. Note that a lower MAPE is better, while a higher R2 is better.*

### 4.3 Incident prediction using traffic flow features

In this section we analyse the impact of leveraging traffic flow features in predicting the incident duration. The purpose is to determine how much of the traffic flow information would improve the prediction, and which would be the best way of making use of this data. The 5 feature data sets described in Section 2 are therefore evaluated using the most performant model determined previously which is the extreme boost decision tree model.

For each incident in the dataset, we construct five descriptions using the following features sets: BFS, FSA, FSB, FSC, FSD (where BFS contains only the baseline incident features, FSA is the most detailed feature set containing all traffic flow information with almost 730 features and FSC and FSD are the most aggregated in terms of traffic flow).

**Results.** The regression results are shown in Fig 8, where blue bar stands for the BFS and grey for BFS + traffic flow features (FSA, FSB, FSC and FSD). The first observation is that using FSA seems to obtain the best MAPE performance among all feature sets (55.28) due to the multitude of feature sets on all road sections, followed closely by FSC (56.06) when we aggregate the traffic flow information of the closest 5 sections to the incident location. But the gain in prediction accuracy of FSA compared to FSC is not significant, especially compared to the higher computational power needed to extract all the traffic flow information from all the subnetwork sections; this aspect makes FSC a more flexible solution for still obtaining a good accuracy. The FSD (with radius dv=500m) seems to obtain the worst MAPE performance among all feature sets (57.93). This indicates that considering the traffic in the immediate 500meters-vicinity of the accident doesn't improve much the prediction compared to considering only the 5 closest road sections. This aspect can be explained by the fact that the majority of our accidents have affected 2 or 3 lanes in a single direction along a road section, and not entire intersections, which happens less often. We make the observation that the 500meters distance radius in FSD has been fixed after a sensitivity analysis conducted around the vicinity area of the accident in which we have tested the model performance with traffic flow from 100 meters up to 600 meters; for our data set, the model presented a 14% improvement when ingesting traffic flow at 500meters distance from the incident location, in comparison with using flow at 100 meters; when ingesting traffic flow information further away from 500 meters, the model accuracy started to decrease by 5%, reason for which we have fixed the surrounding area to 500meters for this feature set.

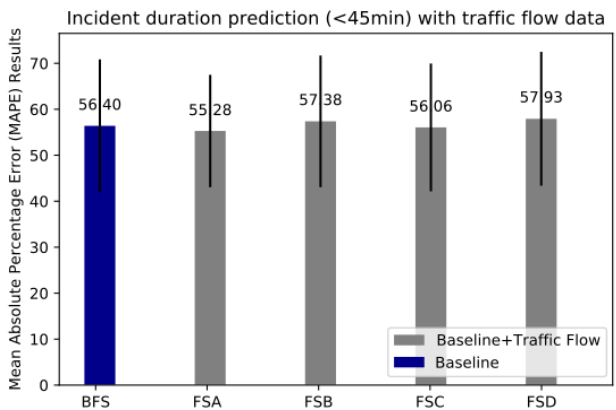

*Fig 8 Incident duration prediction using various feature data sets.*

## 6. Feature importance and discussion

In addition to analysing the predictive performances, we also examine the importance of each feature in the obtained regression tree using the Shapley values previously presented in Eq. 8 (27). Fig 10 demonstrates an example of an incident prediction (estimated at 36.99min), represented with the influence of each feature extracted from Shap values. One can observe that features marked in blue are lowering the incident duration, while features marked in red are increasing the predicted duration. We note that the feature values shown in Fig 10 have been normalized between 0 and 1 just for this example and that the hour-of-day and the X-coordinate of the location are the most important features influencing the prediction outcome. This approach allows us to investigate the contribution of each feature in the final prediction and understand what are the most critical factors that influence the incident duration.

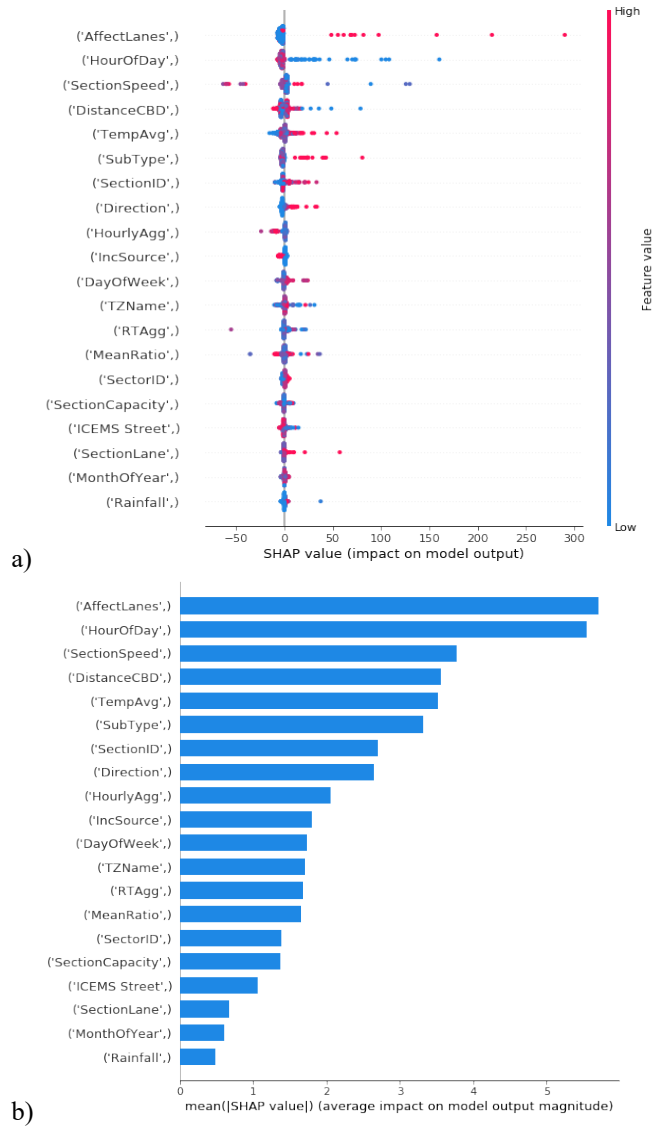

*Fig 9 Feature ranking representation using a) SHAP and b) average feature weights.*

Fig 9a) offers not only an insight into the feature ranking for the predictive method, but also a detailed breakdown of how the values of each feature will increase or decrease the incident duration; for example, each point along a feature analysis represents the predicted incident duration, and the colour of that point reflects the importance of that feature for the overall prediction. The number of affected lanes seems to be the most important which increases the incident duration prediction (see the spread of red dots to the right of the graph which indicated that if the number of affected lanes is high, then the predicted duration will be consistently higher as well – i.e. there is one accident in which the affected number of lanes added almost 280min to the final predicted clearance time). Second most important feature seems to be the hour-of the-day when the accident takes place: if the incident happens at low hours (during the night) the incident duration seems to increase. The thirst most important feature is the section speed which shows that for sections with low speed, the incident will last longer (probably due to accessibility restrictions) while for sections with high speed, the incidents can be reduced by even 60min (see most left dot in the section speed feature of the Shap distribution). Another observation is that incidents happening closer to the CBD take longer to be cleared off. Another important finding is that the aggregated traffic flow data from FSC (see feature "HourlyAggreg") seems to influence the prediction in the following: if traffic flow is higher (basically less congestion) then the incident duration can drop by almost 20 minutes compared to congested cases (when traffic flow drops due to impossibility of cars to move through the affected incident area).

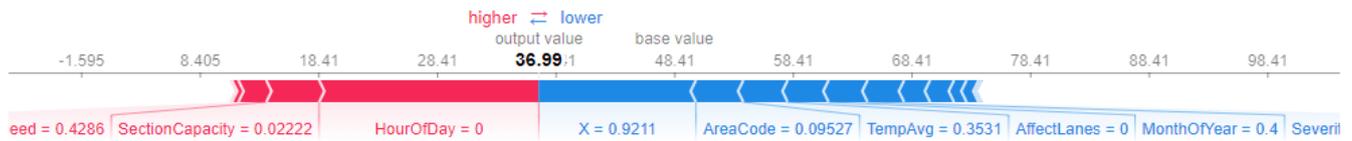

*Fig 10 An example of the importance of features in incident prediction based on SHAP-values.*

*Fig 9*b) shows the summarized ranking of most important features in terms of average SHAP values (mean of all contributions to the predicted incidents from *Fig 9*a)). Once again we observe that the affected lanes and hour of day have the biggest impact overall (average Shap of 6/10 and 5.7/10 respectively), followed by the Section Speed (3.8/10), the Distance from the CBD (3.6/10) and the average temperature (3.6/10). The hourly traffic has a 2.5/10 Shap values, indicating that it can be helpful to improve the incident duration prediction, but it doesn't play the most important role, as revealed previously in Fig 8. The least impactful features seem to be the section lane, month of the year and the Rainfall features, all registering mean Shap values below 1. This is mostly related to weather patterns for Sydney regions which do not experience high extreme temperatures/rainfall throughout the year.

## 7. Conclusions

In this paper we propose a bi-level framework for predicting the incident duration along arterial roads using various models of classification and regression, as well as various method for integrating baseline incident information with available traffic flow information for improving the duration prediction. Firstly, we deployed a classification modelling for predicting if a new incident will be cleared off in less than 45 minutes and we identified GBDT and XGBoost as the best performing classifiers. Secondly, for the short-term incident previously detected, we employed several regression methods to predict the actual incident duration in minutes. We further applied an outlier removal and hyper-parameter tuning and identified that the extreme boost tree method XGBoost trained with MAPE outperforms various other models. To the best of our knowledge this is the first time that such an approach is applied for incident duration prediction. Another contribution of this work is to show the importance and impact of using real-time traffic flow information during the prediction and what would be the best method to integrate it in the feature construction. Our findings indicate that taking the real-time traffic flow of the closest sections as well as one-hour prior to the accident can also improve the results of the prediction.

**Limitations**: While the current results are very promising, a significant limitation of this work is the small size of the dataset used for model training and its long tail distribution; in order to further test the applicability of the proposed framework, we would need to extend the training data set to multiple years for example, while also investigating a more detailed feature generation which would include other useful information such as: proximity to tunnels, motorway exits/entries/loading areas/etc. This would lead to a more correlated feature set to the predicted variable which would be easier to train and validate.

**Perspectives**: The current ongoing work is to extend the approach for predicting as well long-term incident durations (>45min). This, however, proves to be a more difficult task requiring for example natural language processing methods to improve the feature data sets with more information, and even a graph congestion propagation study for estimating the real impact of the accident on the traffic flow. As initially mentioned, predicting incident duration on arterial roads is a very challenging task due to network complexity and high chances of errors when matching corresponding affected sections in the traffic flow feature generation. We plan to extend the current work to motorways incident duration prediction in the city of Sydney while also including more heterogenous incident classes in the learning procedure.

## 8. Acknowledgement

The work presented in this paper is partially funded by the New South Wales Premiere's Innovation Initiative and partially by the ARC LP180100114. The authors of this work are grateful for the work and support of the Transport for New South Wales, Australia. Data61 is funded by the Australian Federal Government through the Commonwealth Scientific and Industrial Research Organization. Marian-Andrei Rizoiu was supported by the Air Force Research Laboratory, under agreement number FA2386-15-1-4018.

## 9. Author contribution

The authors confirm contribution to the paper as follows: data collection, study conception, design, and validation: Dr. Mihaita; data science and model performance: Mr. Liu, Dr. Rizoiu; draft manuscript preparation: all authors. All authors reviewed the results and approved the final version of the manuscript.